\pdfoutput=1

\documentclass[11pt]{article}

\usepackage[final]{acl}

\usepackage{times}
\usepackage{latexsym}

\usepackage[T1]{fontenc}

\usepackage[utf8]{inputenc}

\usepackage{microtype}

\usepackage{inconsolata}
\usepackage{todonotes}
\usepackage{graphicx} 
\usepackage{booktabs} 
\usepackage{multirow} 
\usepackage{amsmath} 
\definecolor{cornflowerblue}{rgb}{0.39, 0.58, 0.93}
\definecolor{coralpink}{rgb}{0.97, 0.51, 0.47}

\usepackage{enumitem}
\usepackage{etoolbox}
\usepackage{subfig}
\usepackage{hyperref}

\newcommand{\logprobs}[0]{{\textsc{LogProbs}}}

\newcommand{\prompting}[0]{{\textsc{Prompting}}}

\title{Log Probabilities Are a Reliable Estimate of Semantic Plausibility \\ in Base and Instruction-Tuned Language Models}

\author{Carina Kauf \\
  Massachusetts Institute of Technology  \\
  \texttt{ckauf@mit.edu} \\\And
  Emmanuele Chersoni \\
  The Hong Kong Polytechnic University \\
  \texttt{emmanuele.chersoni@polyu.edu.hk} \\\AND
   Alessandro Lenci \\
  University of Pisa \\
  \texttt{alessandro.lenci@unipi.it} \\\And
   Evelina Fedorenko \\
  Massachusetts Institute of Technology \\
  \texttt{evelina9@mit.edu} \\\AND
  Anna A. Ivanova\\
  Georgia Tech University\\
  \texttt{a.ivanova@gatech.edu}
  }

\begin{document}

\maketitle

\begin{abstract}

Semantic plausibility (e.g. knowing that ``the actor won the award'' is more likely than ``the actor won the battle'') serves as an effective proxy for general world knowledge. Language models (LMs) capture vast amounts of world knowledge by learning distributional patterns in text, accessible via log probabilities (\logprobs) they assign to plausible vs.~implausible outputs. The new generation of instruction-tuned LMs can now also provide explicit estimates of plausibility via \prompting. Here, we evaluate the effectiveness of \logprobs\ and basic \prompting\ to measure semantic plausibility, both in single-sentence minimal pairs (Experiment 1) and short context-dependent scenarios (Experiment 2). We find that (i) in both base and instruction-tuned LMs, \logprobs\ offers a more reliable measure of semantic plausibility than direct zero-shot \prompting, which yields inconsistent and often poor results; (ii) instruction-tuning generally does not alter the sensitivity of \logprobs\ to semantic plausibility (although sometimes decreases it); (iii) across models, context mostly modulates \logprobs\ in expected ways, as measured by three novel metrics of context-sensitive plausibility and their match to explicit human plausibility judgments. We conclude that, even in the era of prompt-based evaluations, \logprobs\ constitute a useful metric of semantic plausibility, both in base and instruction-tuned LMs.\footnote{Code and data are accessible at \url{https://github.com/carina-kauf/llm-plaus-prob}.}
\end{abstract}

\section{Introduction}

Effective language use heavily relies on general world knowledge.
To determine which sentence is the most appropriate response in a given situation, a language user often needs to establish whether the sentence (e.g., ``The actor won the award'') plausibly describes the world. In NLP, leveraging world knowledge is important both for specific tasks (such as information retrieval) and for general success of a language model during interactions with a user (such as establishing common ground).

Language models (LMs) are well-positioned to acquire many aspects of general world knowledge by capturing distributional patterns in their training data \citep{elazar2022measuring,kang2023impact}. For instance, by observing that ``actor'' occurs more frequently with ``award'' than with ``battle'', the LM might implicitly learn that actors are more likely to win awards than battles. Thus, a simple word-in-context prediction objective can enable an LM to acquire vast amounts of world knowledge. 

We focus on one particular way to assess general world knowledge: estimates of sentence plausibility. Plausible sentences conform with world knowledge whereas implausible sentences violate it; thus, the ability to distinguish plausible and implausible sentences is an indicator of underlying world knowledge capabilities. Plausibility judgments can be tested using both single sentences (e.g., ``The actor won the award'' > ``The actor won the battle'') and setups where plausibility depends on the context of the previous sentences (e.g., ``The girl dressed up as a canary. She had a little beak.'' > ``The girl was cute. She had a little beak.'').

A quantitative metric that has been commonly used to evaluate world knowledge in LMs---including semantic plausibility---are the log probability scores (\logprobs) of the output under the model. \logprobs\ are relatively easy to compute and constitute a direct measure of model behavior (as opposed to more implicit metrics such as decoding probe accuracy; \citealp{li2021implicit, papadimitriou2022classifying}). However, \logprobs\ are sensitive to many different surface-level text properties, such as individual word frequency, output length, and tokenization schemes \cite{holtzman2021surface, salazar-etal-2020-masked, kauf2023better}. Furthermore, distributional patterns are subject to the reporter bias: people typically communicate new or unusual information rather than trivial or commonly known facts \cite{gordon2013reporting}. Thus, the link between \logprobs\ and semantic plausibility is confounded by a variety of factors. 
The most common way to control for confounds influencing \logprobs\ is by leveraging the minimal pairs setup \cite{futrell2019neural, warstadt2020blimp, hu2020systematic,aina2021language,pedinotti2021did,sinha2022language,michaelov2023can, hu2024language, misra2024experimental} and/or quantifying the effects of multiple contributing factors on the resulting score \cite{kauf2023event}.

With the rise of instruction-tuned LMs \cite{chung2022scaling,touvron2023llama,almazrouei2023falcon,jiang2023mistral}, it has become possible to directly evaluate LM capabilities via targeted natural language \prompting\ \cite{li2022probing,blevins2023prompting}. Thus, we ask: is explicitly prompting instruction-tuned LMs for semantic plausibility judgments more effective than using \logprobs-derived plausibility estimates? And how does instruction tuning affect the \logprobs\ estimates themselves? 

On the one hand, \prompting\ might provide a better estimate of plausibility by filtering out influences of extraneous factors not mentioned in the prompt. Furthermore, instruction tuning might diminish the influence of those factors even at the level of \logprobs\ themselves, leading instruction-tuned models to perform better under either metric. On the other hand, initial direct comparisons of \logprobs\ and \prompting\ measures on different linguistic/semantic knowledge datasets revealed that \prompting\ may, in fact, systematically underestimate the model's internal knowledge by requiring the models not only to solve the task, but also to correctly interpret the prompt and to translate their answer into the desired output format \cite{hu2023prompting, hu2024language}. 

As access to \logprobs\ for newer models becomes restricted, it is important to understand what knowledge can be accessed, and what knowledge is inaccessible to the experimenter if \prompting\ is the only way to interact with LMs. In addition, some researchers reported that instruction tuning decreases the utility of raw \logprobs\ in domains such as confidence judgments \cite{tian-etal-2023-just} and prediction of human reading times \cite{kuribayashi2023psychometric}, a change that might or might not be compensated by superior \prompting\ performance and that needs to be acknowledged as the field is shifting toward instruction-tuned LMs.

In this paper, we provide a systematic comparison of semantic plausibility estimates  in instruction-tuned LMs. We test LMs' knowledge of plausibility in single-sentence (Experiment 1) and contextualized scenarios (Experiment 2) and compare implicit (\logprobs-based) and explicit (\prompting-based) plausibility judgments. 
We find that:
\begin{enumerate} 
    \item \logprobs\ \hspace*{-0.2em}, while imperfect, are a more dependable measure of plausibility than naive zero-shot \prompting.
    \item Instruction-tuning does not drastically alter \logprobs-derived plausibility estimates, although in certain cases they might become \textit{less consistent} with human plausibility judgments compared to base model versions.
    \item \logprobs\ can be used to effectively model the \emph{contextual} plausibility of events and replicate key patterns of human plausibility-judgment behaviors in both base and instruction-tuned LMs.
\end{enumerate}

\begin{table*}
\centering
\small
\begin{tabular}{@{}p{2.5cm}cccll@{}}
\toprule
\textbf{Dataset} & \textbf{Plausible?} & \textbf{Possible?} & \textbf{Voice} & \textbf{Example} & \textbf{Source}\\ 
\midrule
\multirow{4}{=}{\begin{minipage}{3cm}
    \textbf{EventsAdapt}\hspace*{0.5em}\includegraphics[width=0.5cm]{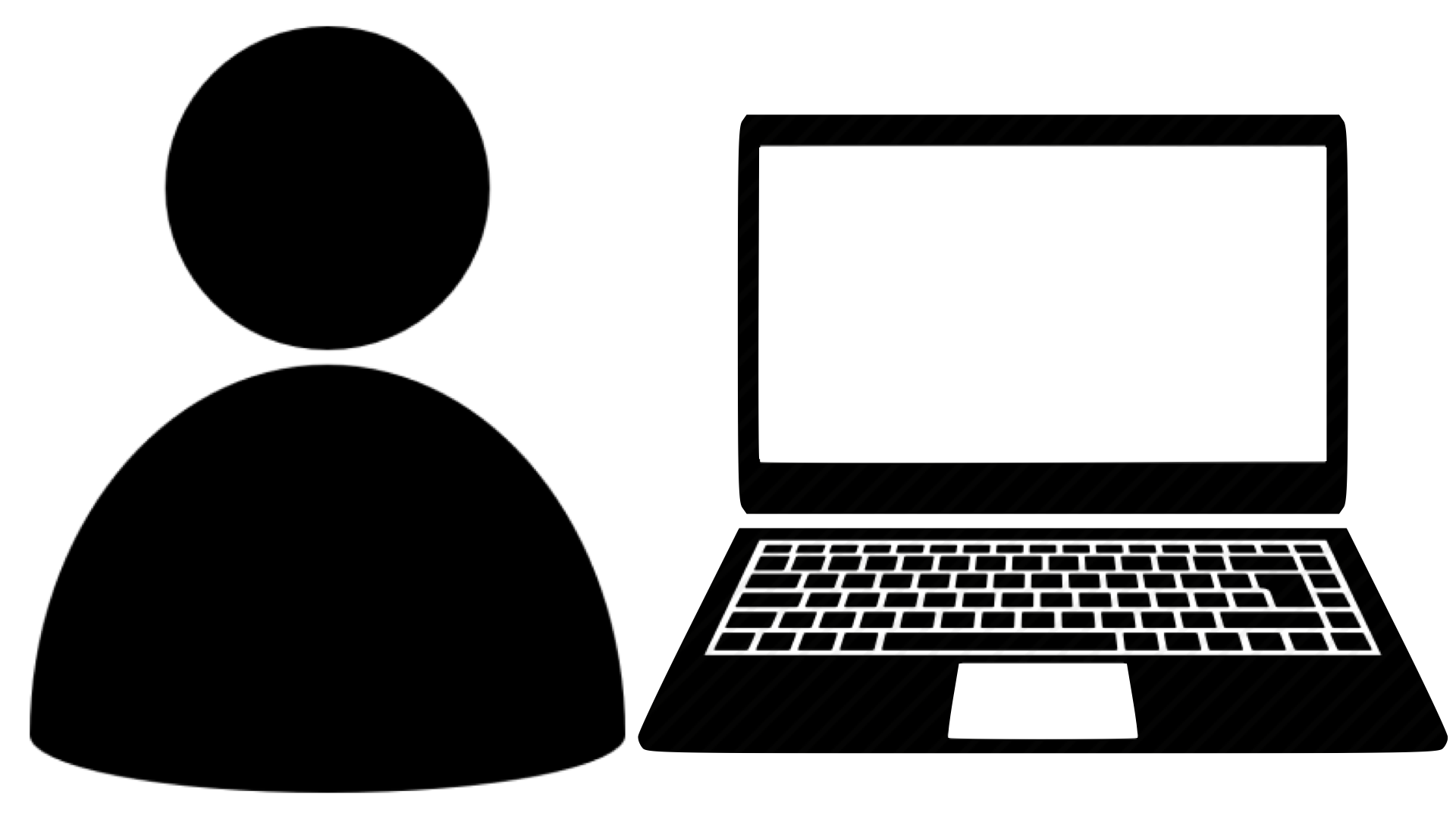}\\
     \textbf{(AI, impossible)}
\end{minipage}} & Yes & Yes & Active & The teacher bought the laptop. & \multirow{8}{*}{\citet{fedorenko2020lack}}\\
 &  &  & Passive & The laptop was bought by the teacher. & \\
& No & No & Active & The laptop bought the teacher. & \\
&  &  & Passive & The teacher was bought by the laptop. & \\
 \cmidrule(lr){1-5}
\multirow{4}{=}{\begin{minipage}{3cm}
    \textbf{EventsAdapt}\hspace*{0.5em}\includegraphics[width=0.5cm]{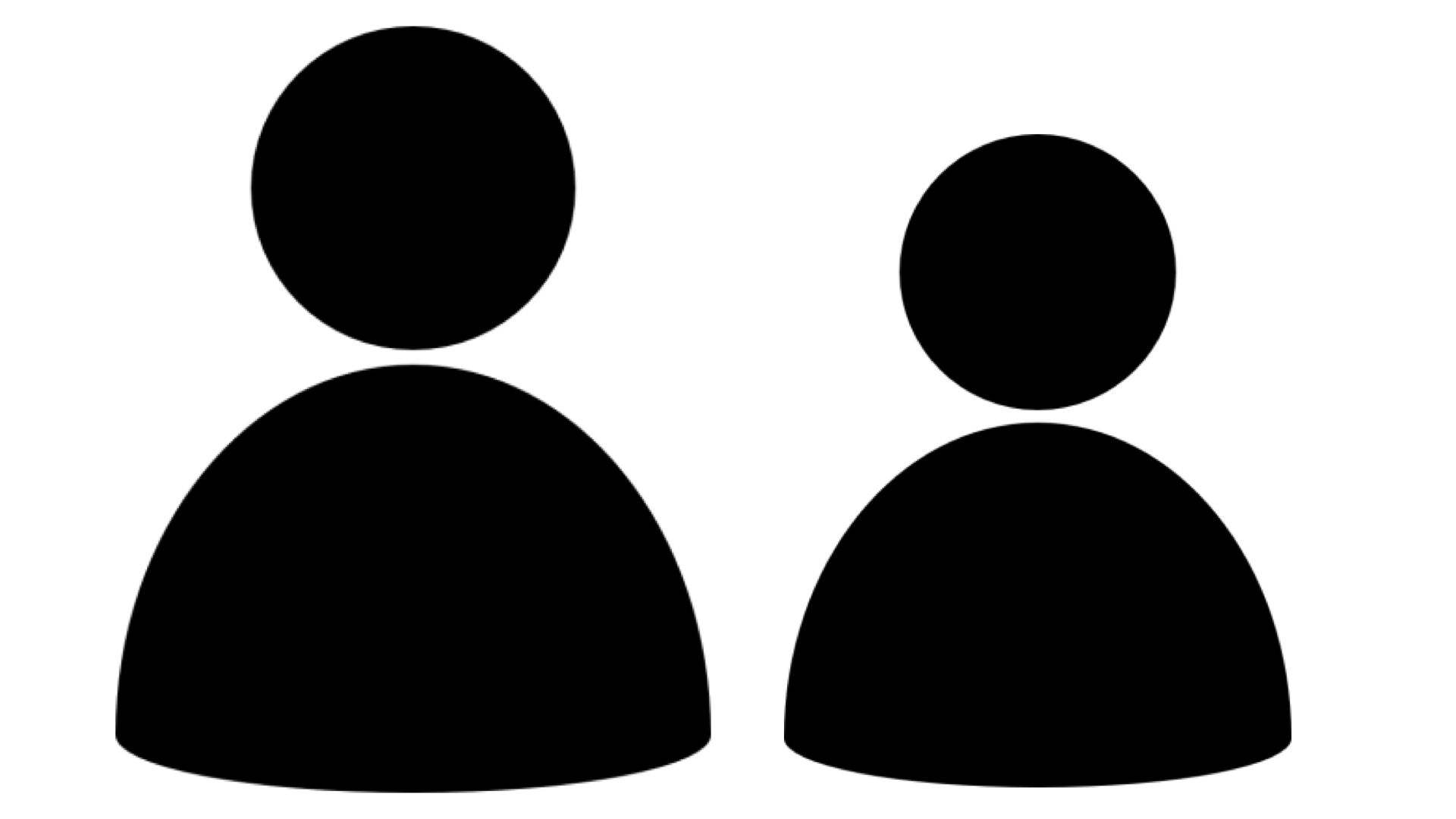}\\
     \textbf{(AA, unlikely)}
\end{minipage}}  & Yes & Yes & Active & The nanny tutored the boy. &  \\
 &  &  & Passive & The boy was tutored by the nanny. & \\
 & No & Yes & Active & The boy tutored the nanny. & \\
 &  & & Passive & The nanny was tutored by the boy. & \\
\midrule
\multirow{2}{=}{\begin{minipage}{3cm}
    \textbf{DTFit}\hspace*{0.5em}\includegraphics[width=0.5cm]{figures/animate-inanimate-laptop.png}\\
     \textbf{(AI, unlikely)}
\end{minipage}}  
    & Yes & Yes & Active & The actor won the award. & \multirow{2}{*}{\citet{Vassallo:2018}}  \\
    & No  & Yes & Active & The actor won the battle. & \\
\bottomrule
\end{tabular}
\setlength{\belowcaptionskip}{-5pt} 
\caption{Example stimuli from the datasets used in Experiment 1. Names in parentheses indicate event participant animacy (AI = animate agent, inanimate patient; AA = animate agent, animate patient) and the plausibility type of the implausible sentences in the dataset (impossible vs.~unlikely).
}
\label{tab1}
\end{table*}

\section{Related Work}

\textbf{Evaluating single-sentence plausibility in LMs.} 
In Experiment 1, we evaluate plausibility estimates for single sentences describing common events (Table \ref{tab1}). 
To evaluate plausibility, scholars traditionally tested NLP models with sentence pairs from psycholinguistic studies that differ for their degree of semantic plausibility (e.g. \textit{The mechanic was checking the brakes} vs. \textit{The journalist was checking the brakes}, from \citealp{bicknell2010effects}): the models' goal is to guess which of the two sentences is the most plausible one \citep{lenci2011composing,tilk2016event,chersoni2016towards,chersoni2019structured,chersoni2021not}.

\citet{pedinotti2021did} and \citet{kauf2023event} specifically tested event plausibility knowledge in non-finetuned LMs. \citet{pedinotti2021did} showed that LMs achieve correlation with human judgments on par with or better than traditional distributional models. \citet{kauf2023event} showed that Transformer-based models retain a considerable amount of event knowledge from textual corpora and vastly outperform the competitor models (i.e., classical distributional models and LSTM baselines). Nevertheless, both studies show LMs' generalization capabilities to novel experimental manipulations of the target sentences are limited and that \logprobs\ are affected by task-irrelevant information, such as the frequency of words within a target sentence.

\vspace*{0.5em}
\noindent
\textbf{Evaluating context-dependent linguistic judgments in LMs.}
In Experiment 2, we evaluate context sensitivity of LM plausibility estimates (Table \ref{tabSocialN400}). Initial work in this domain shows that LMs can modulate their probability estimates to accommodate a previously unlikely target word (e.g., \textit{A peanut falls in love}) following a short licensing context \cite{michaelov2023can, hanna2023language}, results that are consistent with human data \citep{nieuwland2006peanuts, rueschemeyer2015social}. Nevertheless, probability-based judgments of LMs can also be \textit{adversely} influenced by context, for example in cases where the context contains information that is not related to the task (for syntax: e.g., \citealp{sinha2022language}; for factual knowledge: e.g., \citealp{kassner2020negated}).

\vspace*{0.5em}
\noindent
\textbf{Comparing \logprobs\ and \prompting.}
The direct interaction with LMs through natural language prompts is exciting for many reasons, including the ability to run the exact same experiments on models and on humans \citep{lampinen2022can}. Nevertheless, \citet{hu2023prompting, hu2024language} showed that the use of metalinguistic prompts for model evaluation may underestimate their true capabilities. They compared LMs' syntactic/semantic knowledge across four minimal sentence pair datasets and showed that, on average, direct probability measures were a better indicator of these knowledge types than answers to prompts (similar to us, they used \textit{DTFit} as one of their datasets, but their prompts did not explicitly probe the notion of plausibility; thus, we chose to include \textit{DTFit} in this work; see Appendix \S\ref{sec:si-dtfit}, Figure \ref{fig:DTFit-prompting} for a more direct comparison). 

\vspace*{0.5em}
\noindent
\textbf{Evaluating the alignment of instruction-tuned models with humans.} 
Even though instruction-tuning has been claimed to better align the representations of LMs and those computed by the human brain \cite{aw2023instruction}, others show that it does not always help for the alignment at the behavioral level \cite{kuribayashi2023psychometric}. However, the work in this domain is still sparse.

\section{Experiment 1: Single-Sentence Plausibility Judgments}\label{sec:experiment1}

In this section, we test LMs' knowledge of semantic plausibility in \emph{isolated sentences}. We compare  implicit (\logprobs-based) and explicit (\prompting-based) judgments derived from the base and instruction-tuned versions of $3$ state-of-the-art LMs. We also compare LM scores with human plausibility judgments.

\subsection{Datasets}\label{sec:datasets}

\begin{table*}[t!]
\centering
\small
\begin{tabular*}{\textwidth}{@{}p{0.17\textwidth}p{0.80\textwidth}@{}}
\toprule
\textbf{Evaluation type} & \textbf{Example} \\ 
\midrule
\logprobs\ Score & \{{\bf \textcolor[HTML]{469082}{The nanny tutored the boy.}}, {\bf \textcolor[HTML]{b96f7d}{The boy tutored the nanny.}}\} \\[0.2em]
Sentence Choice I & Here are two English sentences: 1) The nanny tutored the boy. 2) The boy tutored the nanny. Which sentence is more plausible? Respond with either 1 or 2 as your answer. Answer: \{{\bf \textcolor[HTML]{469082}{1}}, {\bf \textcolor[HTML]{b96f7d}{2}}\} \\[0.2em]
Sentence Choice II & You are evaluating the plausibility of sentences. A sentence is completely plausible if the situation it describes commonly occurs in the real world. A sentence is completely implausible if the situation it describes never occurs in the real world. Tell me if the following sentence is plausible. The nanny tutored the boy. Respond with either Yes or No as your answer. Answer: \{{\bf \textcolor[HTML]{469082}{Yes}}, {\bf \textcolor[HTML]{b96f7d}{No}}\}\\[0.2em]
Likert Scoring & You will be given a sentence. Your task is to read the sentence and rate how plausible it is. Here is the sentence: The nanny tutored the boy. How plausible is this sentence? Respond with a number on a scale from 1 to 7 as your answer, with 1 meaning "is completely implausible", and 7 meaning "is completely plausible". Answer: 
\{ {\bf \textcolor[HTML]{469082}{7},
        \textcolor[HTML]{598b81}{6},
        \textcolor[HTML]{6c8580}{5},
        \textcolor[HTML]{7f8080}{4},
        \textcolor[HTML]{937a7f}{3},
        \textcolor[HTML]{a6747e}{2},
        \textcolor[HTML]{b96f7d}{1}} \}
\\[0.2em]
Sentence Judgment & Here is a sentence: The nanny tutored the boy. Is this sentence plausible? Respond with either Yes or No as your answer. Answer: \{{\bf \textcolor[HTML]{469082}{Yes}}, {\bf \textcolor[HTML]{b96f7d}{No}}\}\\
\bottomrule
\end{tabular*}
\setlength{\belowcaptionskip}{-10pt} 
\caption{Example evaluation strategies. The prompts are extended and adapted from \citet{hu2023prompting}.}
\label{tab:prompts}
\end{table*}

We use two curated sets of minimal sentence pairs ($n \sim 2000$ overall) adapted from previous studies (for an overview, see Table \ref{tab1}):

\vspace*{0.5em}
\noindent
\textbf{EventsAdapt.}\hspace*{1.2em}The \textit{EventsAdapt} dataset \cite{fedorenko2020lack} is composed of $391$ items, each of which includes (i) a plausible active sentence that describes a transitive event (``The teacher bought the laptop''), (ii) the implausible version of the same sentence, constructed by swapping the noun phrases (``The laptop bought the teacher''), and passive voice alternatives (``The laptop was bought by the teacher'' and ``The teacher was bought by the laptop''). The items fall into one of two categories: \textbf{a)} animate-inanimate items (AI \includegraphics[width=0.5cm]{figures/animate-inanimate-laptop.png}; ``The teacher bought the laptop''), where the swap of the noun phrases leads to impossible sentences; and \textbf{b)} animate-animate ones \includegraphics[width=0.5cm]{figures/animate-animate.png} (AA; ``The nanny tutored the boy''), where role-reversed sentences have milder plausibility violations. Given these differences, we model the two subsets independently.

\vspace*{0.5em}
\noindent
\textbf{DTFit.}\hspace*{1.2em}The \textit{DTFit} dataset \cite{Vassallo:2018} contains $395$ items, each of which includes (i) a plausible active sentence that describes a transitive event (``The actor won the award''); (ii) a less plausible version of the same sentence, constructed by varying the inanimate sentence patient (``The actor won the battle'').

\subsection{Human Plausibility Judgments}\label{sec:human-judgments}

For \textit{DTFit}, participants answered questions of the form ``How common is it for a \{agent\} to \{predicate\} a \{patient\}.'' (e.g. ``How common is it for an actor to win an award?'') on a Likert scale from $1$ (very atypical) to $7$ (very typical) \cite{Vassallo:2018}. For \textit{EventsAdapt}, participants evaluated the extent to which each sentence was ``plausible, i.e., likely to occur in the real world'' on a Likert scale from $1$ (completely implausible) to $7$ (completely plausible) \cite{kauf2023event}. For each sentence, we average judgments across the human participant pool to obtain a single score.

\subsection{Model Plausibility Judgments}

\textbf{Models.}\hspace*{1.2em}We test the base and instruction-tuned versions of three popular autoregressive LMs: {\tt Mistral} \citep{jiang2023mistral}, {\tt Falcon} \citep{almazrouei2023falcon}, and {\tt MPT} \citep{MosaicML2023Introducing}, all of them with $7B$ parameters.

\vspace*{0.5em}
\noindent
\textbf{Metrics.}\hspace*{1.2em}We evaluate LMs using (i) \logprobs\, and (ii) several zero-shot \prompting\ methods (Table \ref{tab:prompts}) \cite{hu2023prompting}. \logprobs\ are calculated as the sum of the log-probabilities of each token $w_i$ in a sentence, conditioned on the preceding sentence tokens $w_{<i}$.
In our main analysis, we evaluate LMs using four natural-language prompts (\textit{Sentence Choice I/II, Likert Scoring} and \textit{Sentence Judgment}; Table \ref{tab:prompts}). These prompts were designed to explicitly query the LMs' knowledge of sentence \textit{plausibility} and use either the same or similar instructions to the task that humans solved (see \S\ref{sec:human-judgments}).\footnote{Note that the \textit{DTFit} dataset was included in \citet{hu2023prompting} where it was evaluated using different models and different prompts. However, they did not explicitly query the models for estimates of \textit{semantic plausibility}, but rather paraphrased the LMs' pretraining task, asking which word ``is most likely to come next''. We include an evaluation of our models on their best-performing prompt for \textit{DTFit} as a supplementary analysis (SI \S\ref{sec:si-dtfit}, Figure \ref{fig:DTFit-prompting}).}
For all prompting methods except \textit{Likert Scoring}, we compare the probabilities that models assign to ground-truth continuations (in {\bf \textcolor[HTML]{469082}{green}}) over implausible continuations (in {\bf \textcolor[HTML]{b96f7d}{red}}). For \textit{Likert Scoring}, we ask models to generate a number from a constrained set of answers, using the {\tt outlines} Python library\footnote{\url{https://github.com/outlines-dev/outlines}}, and compare the generated scores for plausible vs. implausible sentences (the results remain consistent across free vs. constrained generation prompting, see SI \S\ref{sec:si-invariance-prompting}, Figure \ref{fig:free-vs-constrained}). In our main experiment, all prompts are framed using the direct plausibility query ``is plausible''. Supplementary analyses show that this pattern of results remains consistent for alternative queries of plausibility, such as ``makes sense'' (SI \S\ref{sec:si-invariance-prompting}, Figure \ref{fig:makessense-vs-isplausible}) and ``is likely'' (SI \S\ref{sec:si-dtfit}, Figure \ref{fig:DTFit-prompting}).

\vspace*{0.5em}
\noindent
\textbf{Binary accuracy.}\hspace*{1.2em}For each item, we compare the scores/generations of the minimally different plausible and implausible sentence conditions, and compute the binary \textit{accuracy} as the ratio of dataset items in which the LM/the human subject pool assigns a higher score to the plausible vs. the implausible sentence variant. The chance level is 50\% for all benchmarks except \textit{Sentence Judgment}, where, following \citet{hu2023prompting}, we compare the models' propensity to output the ground truth answer in both plausible and implausible settings, leading to a chance performance of 25\%.

\subsection{Results}

\textit{\underline{Result 1:} \logprobs\ results are consistent across models, whereas \prompting\ is hit-or-miss.}

\vspace*{0.2em}
\noindent%
\begin{minipage}{\linewidth}
\makebox[\linewidth]{
  \includegraphics[width=\textwidth]{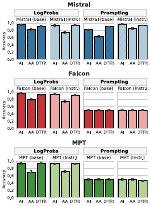}}
\setlength{\abovecaptionskip}{-5pt} 
\captionof{figure}{Results of sentence plausibility judgment performance across models and datasets, using implicit (\logprobs) measures vs. \prompting\ with the best-performing prompt (\textit{Sentence Choice I}). Complete prompting results are shown in SI \S\ref{sec:si-complete-prompting}, Figure \ref{fig:si-mainresult}.}%
\label{fig:main-result}%
\end{minipage}
\vspace*{0.2em}

Across model architectures and plausibility datasets, \logprobs\ are an effective estimate of plausibility knowledge in both base and instruction-tuned LMs (Figure \ref{fig:main-result}). Overall performance patterns across datasets---\textit{DTFit}; \textit{EventsAdapt, AI}; and \textit{EventsAdapt, AA}---are consistent across models, with only minor performance differences. The results are also consistent with prior work \cite{kauf2023event}, showing a performance gap between AI sentences (easier) and AA sentences (harder).

\prompting\ the LMs with our queries, by contrast, yielded inconsistent results. While {\tt Mistral} showed above-chance performance for several prompts, {\tt Falcon} and {\tt MPT} performed at chance level for all prompts tested (for complete prompting results, see  SI \S\ref{sec:si-complete-prompting}, Figure \ref{fig:si-mainresult}). Interestingly, even the base {\tt Mistral} model performed above-chance on some prompts (\textit{Sentence Choice I}), suggesting that model pretraining and/or architecture may be important for the prompt to work in an instruction-tuned model.

Prompts can be tuned to work well for a specific LM and task \citep{qin2021learning,pryzant2023automatic,chen2024prompt}. Even though we do not explore automatic prompt-optimization approaches in this study and instead test variations of the natural-language prompt that humans saw during the experiment (and which people interacting with these models may plausibly use when querying for semantic knowledge in LMs), we observed that certain $\langle$prompt,model$\rangle$ combinations indeed led to improved performance over \logprobs\ (Table \ref{tab:mistral-prompting}). 
Despite this success, however, our comparison critically shows that the same prompt that is effective at tapping into plausibility knowledge in one model class (i.e., \textit{Sentence Choice I} for {\tt Mistral} models) need not be effective in tapping into the same knowledge in other models  (Figures \ref{fig:main-result}, \ref{fig:si-mainresult}). Likewise, we show that the same model that exhibits successful task performance when prompted in a certain way can exhibit poor performance when queried with slight variations on the same prompt (e.g., Table \ref{tab:mistral-prompting}; \citealp[see also][]{sclar2023quantifying}). This brittleness of \prompting-based evaluations stands in contrast to the robustness of the model-agnostic \logprobs-based evaluation scheme of plausibility knowledge in LMs.

\begin{table}[t!]
\centering
\resizebox{0.95\columnwidth}{!}{%
\begin{tabular}{lcc}
\toprule
\textbf{Mistral (EventsAdapt, AA)} & Base & Instruct\\
\midrule
\textit{\logprobs} & \textbf{0.82} (.02) & 0.73 (.03) \\
Sentence Choice I & 0.63 (.02) & \textbf{0.84} (.02)\\
Sentence Choice II & 0.50 (.02) & 0.50 (.02) \\
Likert Scoring & 0.46 (.03) & \textbf{0.61} (.03)\\
Sentence Judgment & 0.14 (.02) & \textbf{0.46} (.03)\\
\bottomrule
\end{tabular}
}
\setlength{\belowcaptionskip}{-10pt} 
\caption{Results of model sentence plausibility judgment performance for {\tt Mistral} on the \textit{EventsAdapt, AA} sentence set shows brittleness of this method. Average performance and standard error around the mean are reported.}
\label{tab:mistral-prompting}
\end{table}

In fact, most of the prompting methods lead to chance-level performance or below-chance performance for most models (Figure \ref{fig:si-mainresult}), even though their log probabilities evidence substantial knowledge about what events are plausible vs. implausible. This result is in line with \citet{hu2023prompting}'s finding of a competence-performance gap when probing models' metalinguistic judgments. 

\vspace*{0.5em}
\noindent
\textit{\underline{Result 2:} \logprobs\ in base and instruction-tuned LMs encode substantial plausibility knowledge but fall short of human performance.}

The \logprobs\ results in Figure \ref{fig:main-result} show that LMs acquire substantial plausibility knowledge from distributional linguistic patterns; all of them performing well above chance on the task. Nevertheless, they also consistently fall short of human performance: On \textit{EventsAdapt (AI, impossible)}, all models were successful in distinguishing plausible and implausible sentences, even though all but one model ({\tt Falcon base}) fell short of human accuracy of $1$ (all Bonferroni-corrected $ps>.05$ except for {\tt Falcon base}: $t=-2.14, p=.02$). On the more challenging \textit{EventsAdapt (AA, unlikely)} subset, all models performed significantly worse than humans in distinguishing AA plausible from implausible events (human accuracy $0.95$; all $ps<.001$). Lastly, the high task performance on \textit{DTFit} shows that LMs can distinguish plausible and implausible AI event descriptions even when low-level distributional cues (like selectional preference restrictions) cannot be used to distinguish the minimal pairs. Despite this success, all models still fall short of human performance of $0.99$ for this dataset at $ps<.001$.

\vspace*{0.5em}
\noindent
\textit{\underline{Result 3:} Instruction tuning can worsen \logprobs\ sensitivity to semantic plausibility.}
\begin{table}[t!]
\resizebox{\columnwidth}{!}{%
\begin{tabular}{lcccccc}
\toprule
 & \multicolumn{2}{c}{{\tt Mistral}} &
 \multicolumn{2}{c}{{\tt Falcon}} &
 \multicolumn{2}{c}{{\tt MPT}}\\
\cmidrule(lr){2-3} \cmidrule(lr){4-5} \cmidrule(lr){6-7}
 & Base & Instruct & Base & Instruct & Base & Instruct\\
\midrule
AA\hspace*{0.5em}\includegraphics[width=0.5cm]{figures/animate-animate.png} & \textbf{0.82**} & 0.73 & \textbf{0.79} & 0.74 & \textbf{0.71} & \textbf{0.71}\\
AI\hspace*{0.5em}\includegraphics[width=0.5cm]{figures/animate-inanimate-laptop.png}& \textbf{0.95} & 0.93 & \textbf{0.97*} & 0.94 & \textbf{0.93} & \textbf{0.93}\\
DTFit\hspace*{0.5em}\includegraphics[width=0.5cm]{figures/animate-inanimate-laptop.png} & 0.91 & \textbf{0.93*} & \textbf{0.92} & 0.91 & \textbf{0.93} & \textbf{0.93}\\
\bottomrule
\end{tabular}
}
\setlength{\belowcaptionskip}{-10pt} 
\caption{\logprobs\ results across models and datasets. Significant differences from dependent t-tests between Base and Instruct models are marked with asterisks ($p<.05$: *; $p<.01$: **).}
\label{tab:ll-baseVsInstruct}
\end{table}

\begin{table*}[t!]
\centering
\small
\begin{tabular}{@{}lllcl@{}}
\toprule
& \multirow{2}{*}{} & \multicolumn{3}{c}{\textbf{Target sentence}} \\
\cmidrule(lr){3-5}
\textbf{Condition} & \textbf{Context sentence (optional)} & \textbf{Prefix}  & \textbf{Tgt. word} & \textbf{Spill-over region}\\ 
\midrule
{\tt Control} & The kids were looking at a canary in the pet store. & The bird had a little & beak & and a bright yellow tail. \\
{\tt SemAnom} & Anna was definitely a very cute child. & The girl had a little & beak & and a bright yellow tail.  \\
{\tt Critical}  & The girl dressed up as a canary for Halloween. & The girl had a little & beak & and a bright yellow tail.  \\
\bottomrule
\end{tabular}
\setlength{\belowcaptionskip}{-10pt}
\caption{\normalsize Sentence manipulations in the dataset by \citet{jouravlev2019tracking}. Tgt. -- Target.}
\label{tabSocialN400}
\end{table*}

Next, we zoom in on the comparison of \logprobs\ derived from base vs. instruction-tuned variants of the same model. Because instruction tuning constrains model behaviors to align with human-desired response characteristics \citep{zhang2023instruction,chia2023instructeval}, it is reasonable to assume that the models' learned probability distributions align better with human expectations of plausible sequences than the base variant, which might be more susceptible to the reporting bias in textual corpora \cite{gordon2013reporting}.

\vspace*{0.5em}
\noindent%
\begin{minipage}{\linewidth}
\makebox[\linewidth]{
  \includegraphics[width=\textwidth]{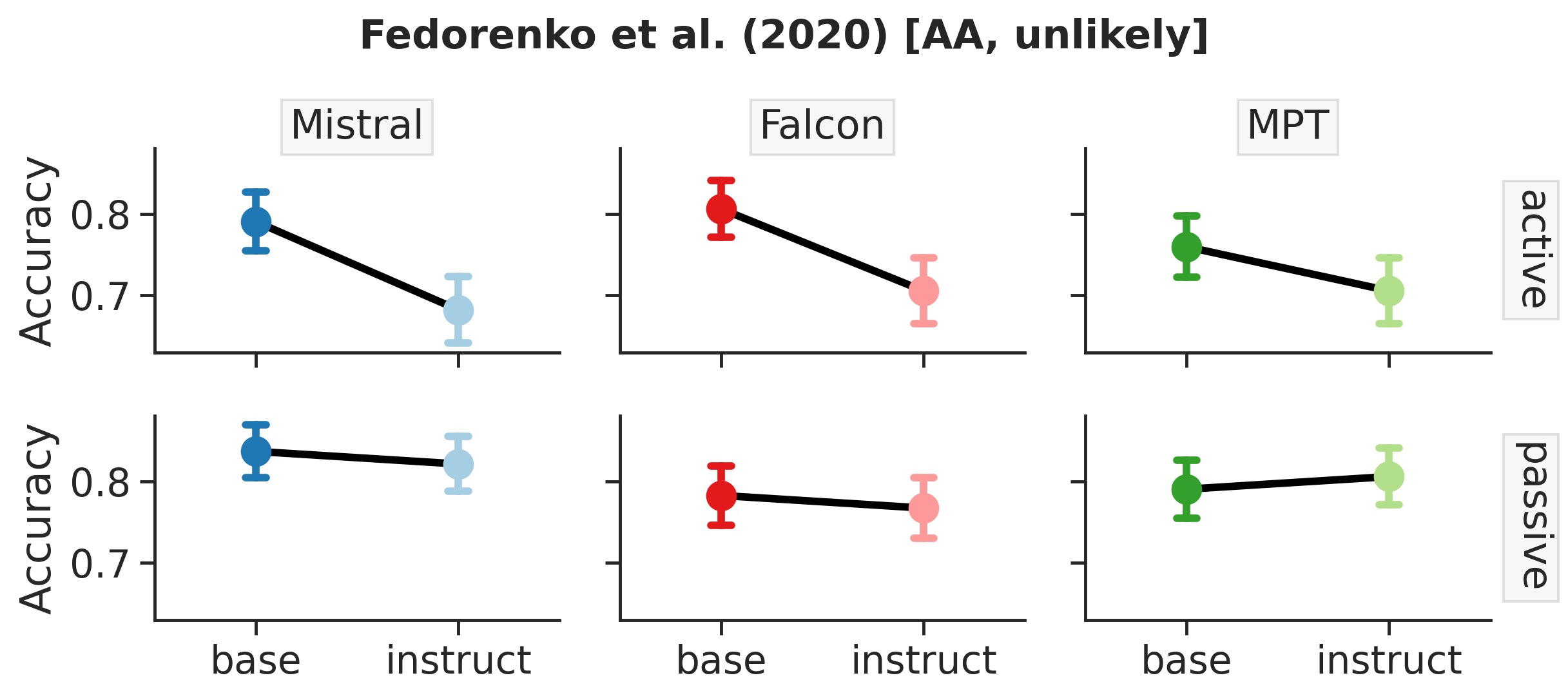}}
\setlength{\abovecaptionskip}{-5pt}
\captionof{figure}{Base vs. instruct model performance in active and passive sentence pairs}%
\label{fig:activeVsPassive}
\end{minipage}
\vspace*{0.2em}

A comparative analysis of the results of base and instruction-tuned model variants across architectures reveals no beneficial effect of instruction-tuning for gauging event plausibility through \logprobs\ measurements: In all but one instance do instruction-tuned models perform similar or even slightly worse than their corresponding base model (Table \ref{tab:ll-baseVsInstruct}). Interestingly, the gap is most noticeable for the most challenging dataset, \textit{EventsAdapt (AA, unlikely)}. An investigation of this difference shows that certain low-level features of the input may disproportionately affect the \logprobs\ that instruction-tuned models assign to word sequences: much of the performance difference is due to the instruction-tuned models' worse performance in discerning plausible and implausible active-voice sentences (see Figure \ref{fig:activeVsPassive}). We quantify these effects by modeling accuracy in a generalized linear mixed-effects model (GLMM). The model uses LLM model class ({\tt Mistral}, {\tt Falcon}, {\tt MPT}), model version ({\tt base}, {\tt instruct}), and voice (active, passive) as fixed effects, and items as random effects (for further GLMM model specification, see SI \S\ref{sec:glmm}). We observed a main effect of model version ($\beta = 0.36, p < .001$) and a significant interaction between model version and active vs. passive voice  ($\beta = -0.37, p < .01$).

This variance highlights the fact that even though direct measurements of model-derived string \logprobs\ in many cases encode task-relevant information (e.g., modeling of grammaticality, \citet{warstadt2020blimp}, of N400 effects, \citet{michaelov2020how}, etc.), they are additionally influenced by low-level features of the input \cite{pedinotti2021did,kauf2023event}.

\section{Experiment 2: Context-Dependent Plausibility Judgments}

Experiment 1 has shown that \logprobs\ are a reliable, albeit imperfect, metric for probing the plausibility of isolated sentences in LMs in both base and instruction-tuned models, whereas \prompting\ measures are brittle and can underestimate the degree of semantic plausibility knowledge LMs encode. However, most of the time, LMs (and humans) do not process sentences in isolation, but rather as part of a larger context. In Experiment 2, we therefore compare LM judgments of semantic plausibility in \textit{short context-dependent scenarios}. Given the success of \logprobs\ over \prompting\ in Experiment 1, we focus on comparing \logprobs\ as measures of context-dependent sentence plausibility in base and instruction-tuned models. Specifically, we compare how the presence of (i) supporting or (ii) non-supporting but related single-sentence contexts modulates the LMs' \logprobs\ judgments. Additionally, we report results for the exact replication of the human study using \textit{Sentence Judgment} prompts.

\subsection{Dataset}

To test the sensitivity of the LM plausibility judgments to discourse context effects, we use a dataset from language neuroscience, collected by \citet{jouravlev2019tracking}. This dataset includes $100$ items in three experimental conditions: a control condition ({\tt Control)}, in which the target sentence describes a plausible situation and the (optional) context sentence adds extra information; a semantically anomalous condition ({\tt SemAnom}), in which the target sentence describes an implausible situation and the context sentence does not provide licensing information; and a critical condition ({\tt Critical}), which shares the same target sentence with {\tt SemAnom}, but here, the context sentence makes it plausible (see the examples in Table \ref{tabSocialN400}).
\subsection{Metrics}

We introduce three critical metrics to evaluate the models' context-aware plausibility judgments:
\newline

\noindent
\textbf{General Plausibility.} This metric measures the propensity of models to assign a higher probability to plausible sentences than to minimally different implausible sentence variants when no influencing context is present (similar to \S\ref{sec:experiment1}). For every dataset item, we assign a model a hit in case
\begin{equation*}
    P(\mathrm{target}_{\tt Contr.}) > P(\mathrm{target}_{\tt Crit.}).
    \label{gen-plaus}
\end{equation*}

\vspace*{0.5em}
\noindent
\textbf{Context-Dependent Plausibility.} This metric measures the ability of models to increase the probability they assign to an \textit{a priori} implausible sentence in the presence of a licensing context. For every dataset item, we assign a model a hit in case
\begin{equation*}
    P(\mathrm{target}_{{\tt Crit.}} | \mathrm{context}_{\tt Crit.}) > P(\mathrm{target}_{\tt Crit.}).
    \label{cd-plaus}
\end{equation*}

\vspace*{0.5em}
\noindent
\textbf{Context Sensitivity.} This metric measures the models' ability to \textit{selectively} update sentence probabilities. For every dataset item, we assign a model a hit in case
\begin{equation*}
    \begin{split}
        P(\mathrm{target}_{\tt Crit.}& | \mathrm{context}_{\tt Crit.}) > \\
        &P(\mathrm{target}_{\tt Crit.} | \mathrm{context}_{\tt Anom.}).
    \end{split}
    \label{context-sens}
\end{equation*}

\subsection{Target region}
For each metric, we evaluate model performance through the likelihood they assign either (i) a critical word within the target sentence or (ii) the target sentence as a whole. If a critical word consists of multiple tokens, we use the sum of the log likelihood scores of the word tokens. Whereas \textit{critical/target word} likelihoods measure the ability of models to detect a contextually unexpected linguistic event, \textit{target sentence} likelihood measures investigate whether implausibility is reliably reflected in the probability the models assign to tokens after encountering a semantically anomalous item, as well. This is because token likelihoods for plausible and implausible sentences are identical until the first contextually unlicensed word appears.

\subsection{Results}

\begin{figure*}
    \centering
    \includegraphics[width=\textwidth]{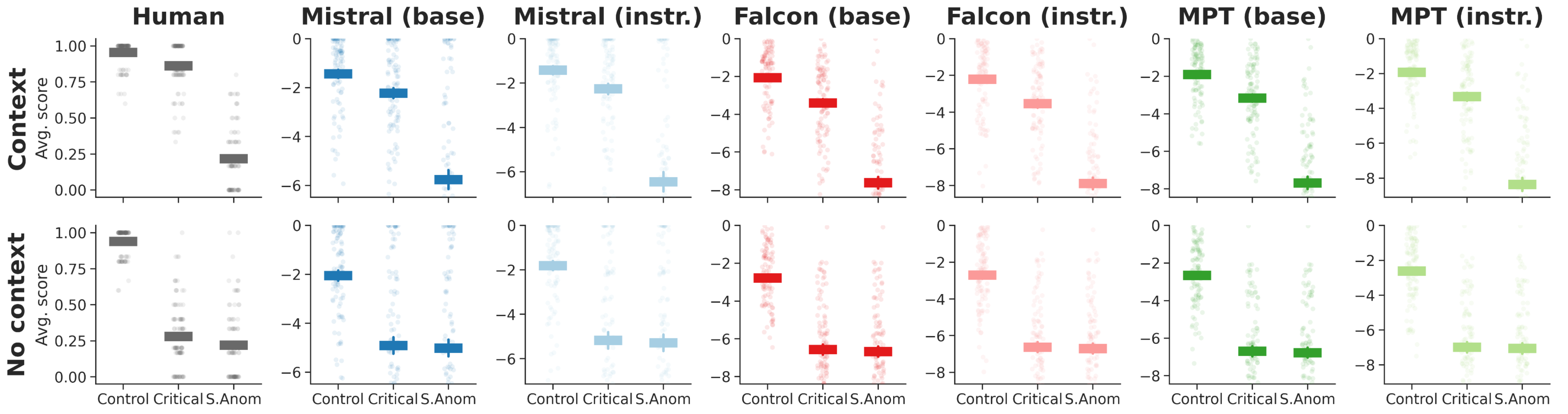}
    \setlength{\belowcaptionskip}{-5pt}
    \setlength{\abovecaptionskip}{-5pt}
    \caption{Target word \logprobs\ replicate patterns of human sentence sensibility judgments. Human data from \citet{jouravlev2019tracking}. Bars indicate average plausibility of sentences (Human) and average target word log likelihoods (LMs). Dots represent individual sentence scores (averaged across the participant pool for Human).} 
    \label{fig:jouravlev2019}
\end{figure*}

\begin{figure*}[h!]
    \centering
    \includegraphics[width=\textwidth]{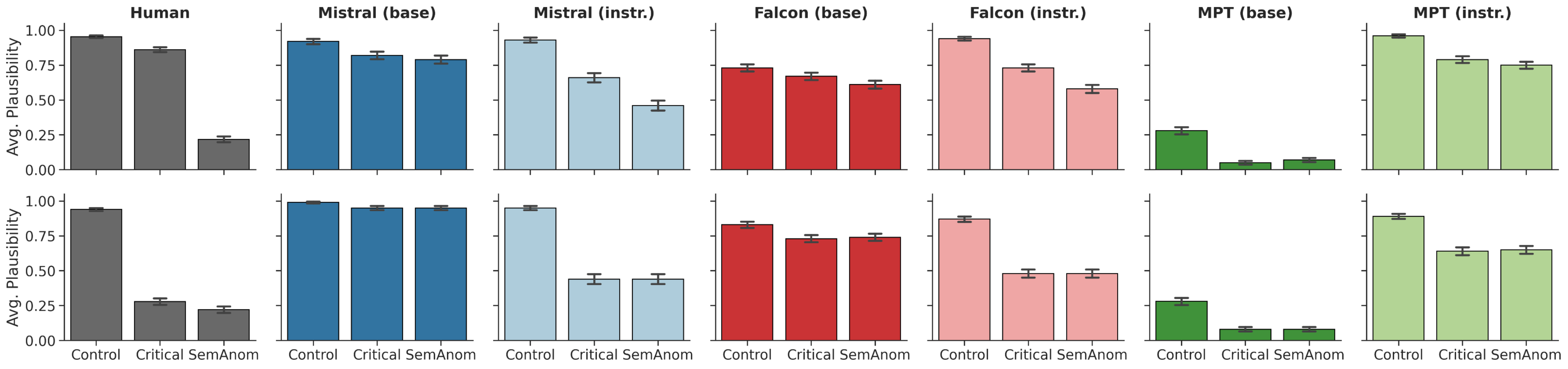}
    \setlength{\belowcaptionskip}{-5pt} 
    \setlength{\abovecaptionskip}{-5pt} 
    \caption{Replicating the sensibility-judgment task in LMs using prompting via the adjusted \textit{Sentence Judgment} prompt in \S\ref{sec:si-jouralev-prompting}. Human data from \citet{jouravlev2019tracking}. We use a barplot to visually set apart this prompt-based comparison vs. \logprobs-based ones in Figures \ref{fig:jouravlev2019}, \ref{fig:jouravlev2019-sentence-ll}.} 
    \label{fig:jouravlev2019-judge}
\end{figure*}

\begin{table}[b!]
\resizebox{\columnwidth}{!}{%
\centering
\footnotesize
\begin{tabular}{lcccccc}
\toprule
 & \multicolumn{2}{c}{\hyperref[gen-plaus]{\textbf{Gen. Plaus.}}} &
 \multicolumn{2}{c}{\hyperref[cd-plaus]{\textbf{Context. Plaus.}}} &
 \multicolumn{2}{c}{\hyperref[context-sens]{\textbf{Context Sens.}}}\\
\cmidrule(lr){2-3} \cmidrule(lr){4-5} \cmidrule(lr){6-7}
 & Word & Sent. & Word & Sent. & Word & Sent.\\
\midrule
{Mistral (base)} & 0.90 & \textbf{0.93} & 0.93 & \textbf{1.00} & \textbf{0.97} & 0.79 \\
{Mistral (instr)} & \textbf{0.97} & 0.90 & 0.93 & \textbf{1.00} & \textbf{0.90} & 0.84 \\
\midrule
{Falcon (base)} & \textbf{0.96} & 0.94 & \textbf{0.93} & 0.92 & \textbf{0.98} & 0.79 \\
{Falcon (instr)} & \textbf{0.98} & 0.91 & \textbf{0.95} & \textbf{0.95} & \textbf{0.96} & 0.77 \\
\midrule
{MPT (base)} & \textbf{0.96} & 0.93 & 0.95 & \textbf{1.00} & \textbf{0.99} & 0.76 \\
{MPT (instr)} & \textbf{0.94} & 0.93 & 0.93 & \textbf{1.00} & \textbf{0.95} & 0.80 \\
\bottomrule
\end{tabular}%
}
\caption{\logprobs\ results for Expt 2. Gen.--General; Context.--Context-Dependent; Plaus.--Plausibility; Sens.--Sensitivity; Word/Sent.--scores for target word/sentence.}
\label{tab:ll-socialn400}
\end{table}

\textit{\underline{Result 1:} Across models, context successfully modulates the \logprobs\ of (im)plausible target words, but not (im)plausible target sentences.}\label{sec:context-dependent-word-vs-sent}\vspace*{0.3em}

When comparing target word vs. target sentence \logprobs, a clear trend emerges: all models demonstrate consistently high performance (around $95$\%) across all metrics when comparing the probabilities of target words (Table \ref{tab:ll-socialn400}, \textit{Word} columns); at the same time, when using the likelihoods they assign to sentences as an indicator of event plausibility knowledge, \logprobs\ plausibility judgments fail to reliably pass the sensitivity criterion. In particular, even though almost all LMs are able to distinguish plausible and implausible sentences (\hyperref[gen-plaus]{General Plausibility}, similar to \S\ref{sec:experiment1}); and are able to modulate the probability they assign an unexpected sentence in the presence of licensing context, they fail to update the sentence probabilities \textit{selectively} (this is evidenced by the substantial drop in performance for the \hyperref[cd-plaus]{Context Sensitivity} metric across LMs). This pattern suggests that while a semantically licensing context assists the models in up-weighing the probability of an otherwise implausible target word/event description (see \hyperref[cd-plaus]{Context-Dependent Plausibility}; in line with \citealp{michaelov2023can}), contextual \textit{implausibility} is not reliably reflected in LMs' sentence likelihoods. In particular, once an unexpected target word has been encountered (which the LMs are able to discern, see \hyperref[cd-plaus]{Context Sensitivity}, \textit{Word} columns), the LMs appear to quickly adjust the predictions in the post-target region, in some cases assigning even higher probabilities to post-target words than in the {\tt Critical} condition, with the consequence that the scores for anomalous sentences and contextually-licensed ones differ less significantly at the sentence level.
This suggests that a semantically-licensing context helps a model in predicting an otherwise anomalous word, but the global probability of the target sentence is less affected by the specific context.

\vspace*{0.5em}
\noindent
\textit{\underline{Result 2:} Context-modulated \logprobs\ align with human contextual judgment patterns.}\label{sec:context-dependent-human}\vspace*{0.3em}

Finally, we investigate how contextual plausibility judgments correspond to human behavior for the same stimuli. We focus on the sensibility-judgment task, in which participants were asked to decide (i) if a target sentence made sense to them within the provided context, or (ii) if it made sense to another person who did not have access to the context sentence \cite{jouravlev2019tracking}. Here, we model this dataset in a `single-participant setting', by exposing the LMs to the full items and comparing the log probabilities assigned to the target words in the three experimental conditions, with or without licensing context. Across models, we see a remarkable match between human- and model-derived plausibility scores, both in the isolated sentence and the contextualized setup (Figure \ref{fig:jouravlev2019}; for supporting statistical analyses see SI \S\ref{sec:stats-socialN400}, Tables \ref{tab:stats-socialn400-ll}/\ref{tab:stats-socialn400-ll-ttest}). 

\logprobs\ again provide a better fit to human data than \prompting\ (Figures \ref{fig:jouravlev2019}, \ref{fig:jouravlev2019-judge}; SI \S \ref{sec:stats-socialN400}, Tables \ref{tab:stats-socialn400-ll}/\ref{tab:stats-socialn400-ll-ttest} vs. Tables \ref{tab:stats-socialn400-judge}/\ref{tab:stats-socialn400-judge-ttest}), although it is interesting to observe that the prompting results for Instruct models matched the human behavioral patterns qualitatively (see also SI \S\ref{sec:si-jouralev-prompting}, \S\ref{sec:si-jouralev-sentence-ll}).

\section{Conclusion}

Overall, we show that, for both base and instruction-tuned models, \logprobs\ remain a more reliable measure of semantic plausibility than naive zero-shot \prompting. This is true in scenarios that evaluate both isolated and context-dependent sentence plausibility.
Even though instruction-tuning has been claimed to align LMs and human brain representations \citep{aw2023instruction}, other studies show that it does not always help for the alignment at the behavioral level \citep{kuribayashi2023psychometric}. Our results show that the base \logprobs\ estimates for simple world knowledge scenarios do not drastically change as a result of instruction tuning, showing approximately the same amount of implicitly encoded information as representation derived from next-word prediction. In some cases, however, instruction tuning can lead to \textit{less} alignment of \logprobs\ to human plausibility judgments than those of base model versions.

Concerning LMs' sensitivity to sentence context, we observe that by using \logprobs\ at the level of the target word, all the models perform around $90$\% with respect to the ground truth and are well aligned to human judgement patterns. However, when using sentence-level \logprobs\, we notice that the models have the tendency to ``re-balance'' the log likelihoods after processing an unexpected word, with the consequence that  semantically anomalous sentences and contextually-licensed ones become harder to distinguish.

Although it is possible that model- and task-specific prompts will outperform raw \logprobs\ as a way to estimate sentence plausibility, our work highlights that \logprobs\ are an easy, zero-shot way to assess LMs' implicit knowledge. Thus, getting a raw \logprobs\ estimate of model performance can provide an initial estimate of whether or not custom prompt-based solutions can be successful or---in some cases---obviate the need for prompt tuning altogether.

\section*{Limitations}

A first, obvious limitation of this work is that it has been conducted on English datasets, so we cannot be sure that our findings on LMs and event knowledge would generalize to other languages.

Second, even though our prompting setup mimics that of humans, it differs in substantial ways. For example, whereas we ask LMs to evaluate sentences in isolation, participants assign scores within the context of the full experiment, having access to their answer history.

Lastly, we only focused on LMs up to $7$ billion parameters, due to the limit of our computational resources, and we only used three representative models in their Base and in their Instruct version. It is possible that with larger and more powerful models the performance will improve and the existing gap with human performance on distinguishing plausible vs. implausible sentences will be closed (cf. \citealp{kauf2023event}).

\section*{Ethical Considerations}
Our work aims to better understand and characterize the capacities of models, and contributes to work highlighting the importance of open access to model representations. Our work shows that LM pre-training distills a wealth of world knowledge into the models' weights, but cannot guarantee the consistency of these representations with human world knowledge. Consequently, LMs should not be expected to generate statements that are consistent with human world knowledge. General ethical concerns about LMs and their impact on human life, especially as they become more and more integrated into people's everyday lives, also apply to our work.

\section*{Acknowledgements}
CK and this work was partially supported by the MIT Quest for Intelligence. EC was supported by a GRF grant from the Research Grants Council of the Hong Kong Special Administrative Region, China (Project No. PolyU 15612222). We would like to thank the three anonymous reviewers for their constructive comments and suggestions.

\bibliography{custom}

\newpage
\appendix

\section*{Supplementary Information}

\section{Complete prompting results}\label{sec:si-complete-prompting}

\begin{figure*}%
    \centering
    \subfloat[\centering EventsAdapt (AA, unlikely)]{{\includegraphics[width=0.33\textwidth]{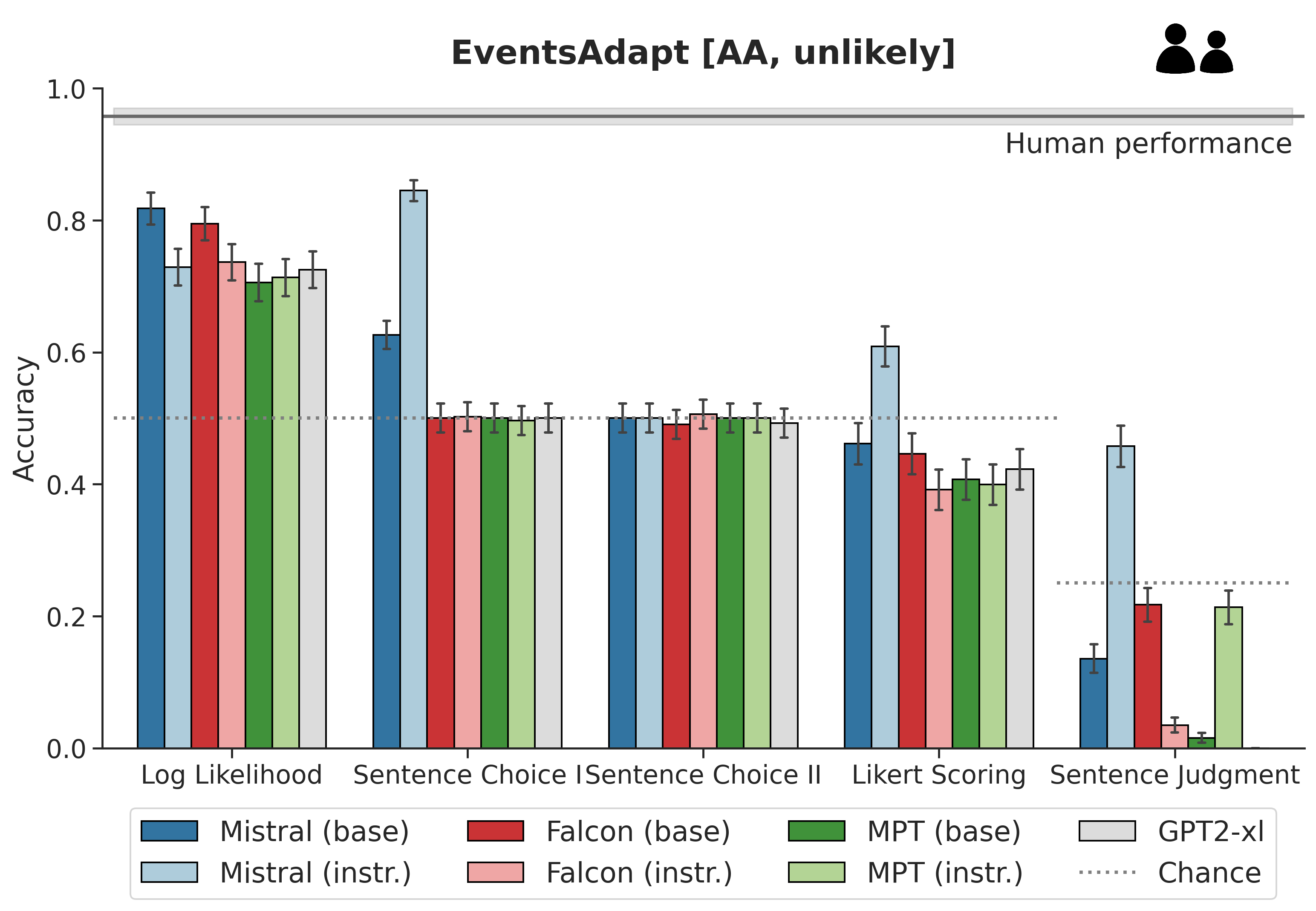} }}%
    \subfloat[\centering DTFit (AI, unlikely)]{{\includegraphics[width=0.33\textwidth]{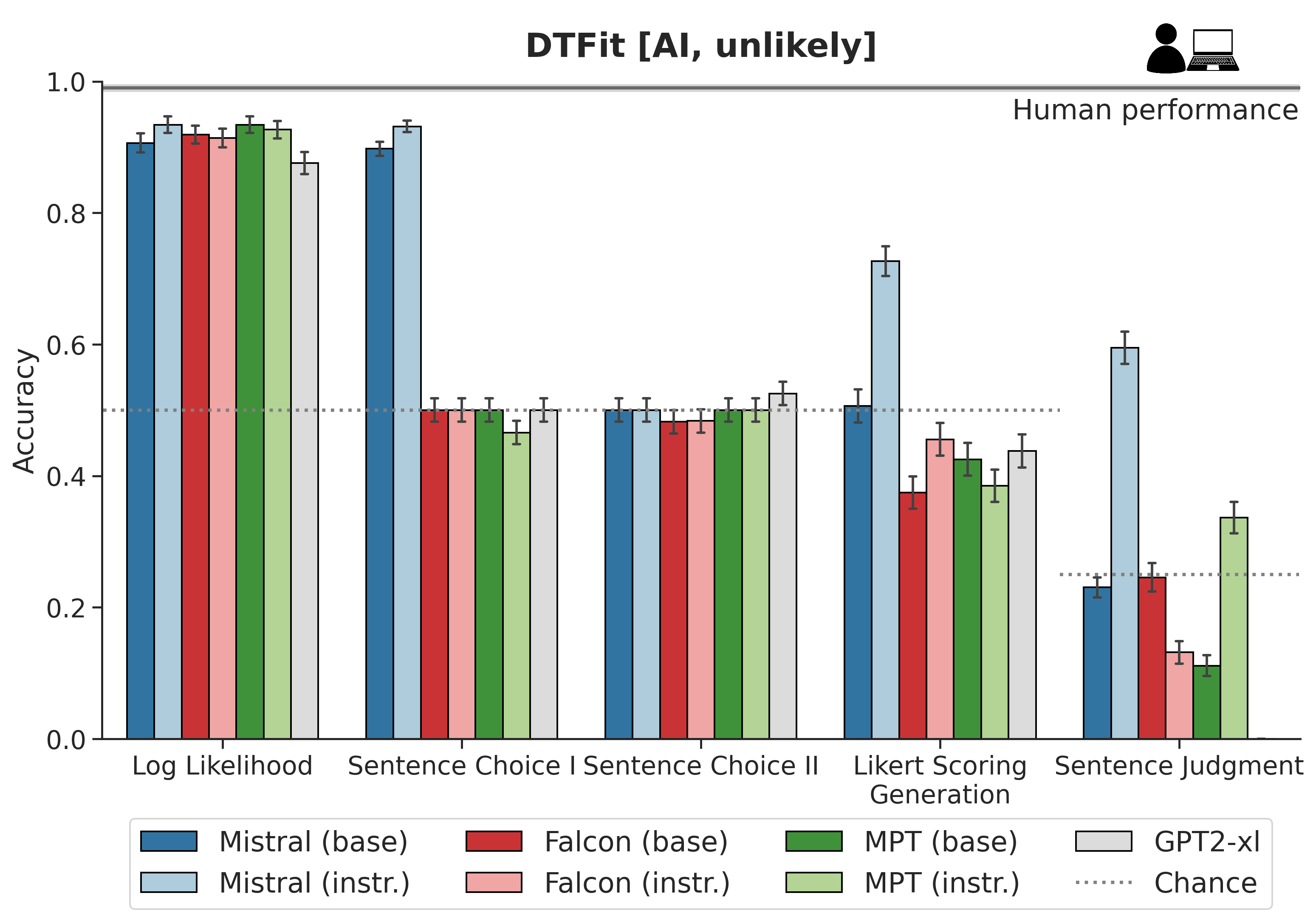} }}%
    \subfloat[\centering EventsAdapt (AI, impossible)]{{\includegraphics[width=0.33\textwidth]{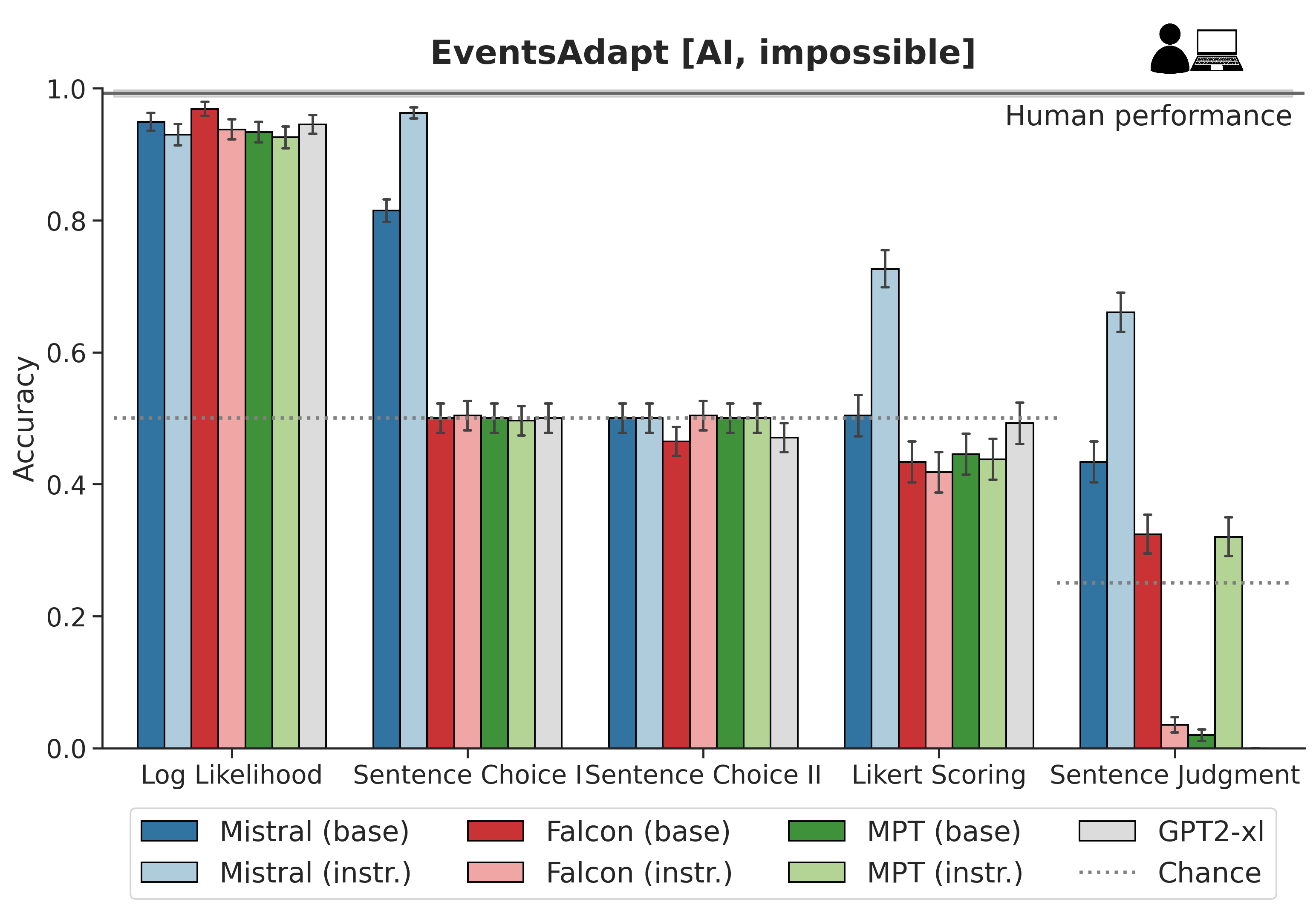} }}%
    \caption{Results of implicit vs. explicit plausibility judgment performance experiments}%
    \label{fig:si-mainresult}%
\end{figure*}

Figure \ref{fig:si-mainresult} shows the complete prompting results across datasets, models and prompts.

\section{Additional prompting results for DTFit}\label{sec:si-dtfit}

\noindent%
\begin{minipage}{\linewidth}
\centering
\small
\begin{tabular*}{\textwidth}{@{\extracolsep{\fill}}p{0.20\textwidth}p{0.75\textwidth}}
\toprule
\textbf{Prompt} & \textbf{Example} \\ 
\midrule
Word Comparison  &  What word is most likely to come next in the following sentence (award, or battle)? The actor won the \{{\bf \textcolor[HTML]{469082}{award}}, {\bf \textcolor[HTML]{b96f7d}{battle}}\} \\
\bottomrule
\end{tabular*}
\captionof{table}{Additional prompt used for \citet{Vassallo:2018} evaluation in Figure \ref{fig:DTFit-prompting}. This prompt is the best-performing prompt for this dataset in \citet{hu2023prompting}.}
\label{tab:additional-prompts}
\end{minipage}
\vspace*{1em}
\noindent%
\begin{minipage}{\linewidth}
\centering
\makebox[\linewidth]{
\includegraphics[width=\textwidth]{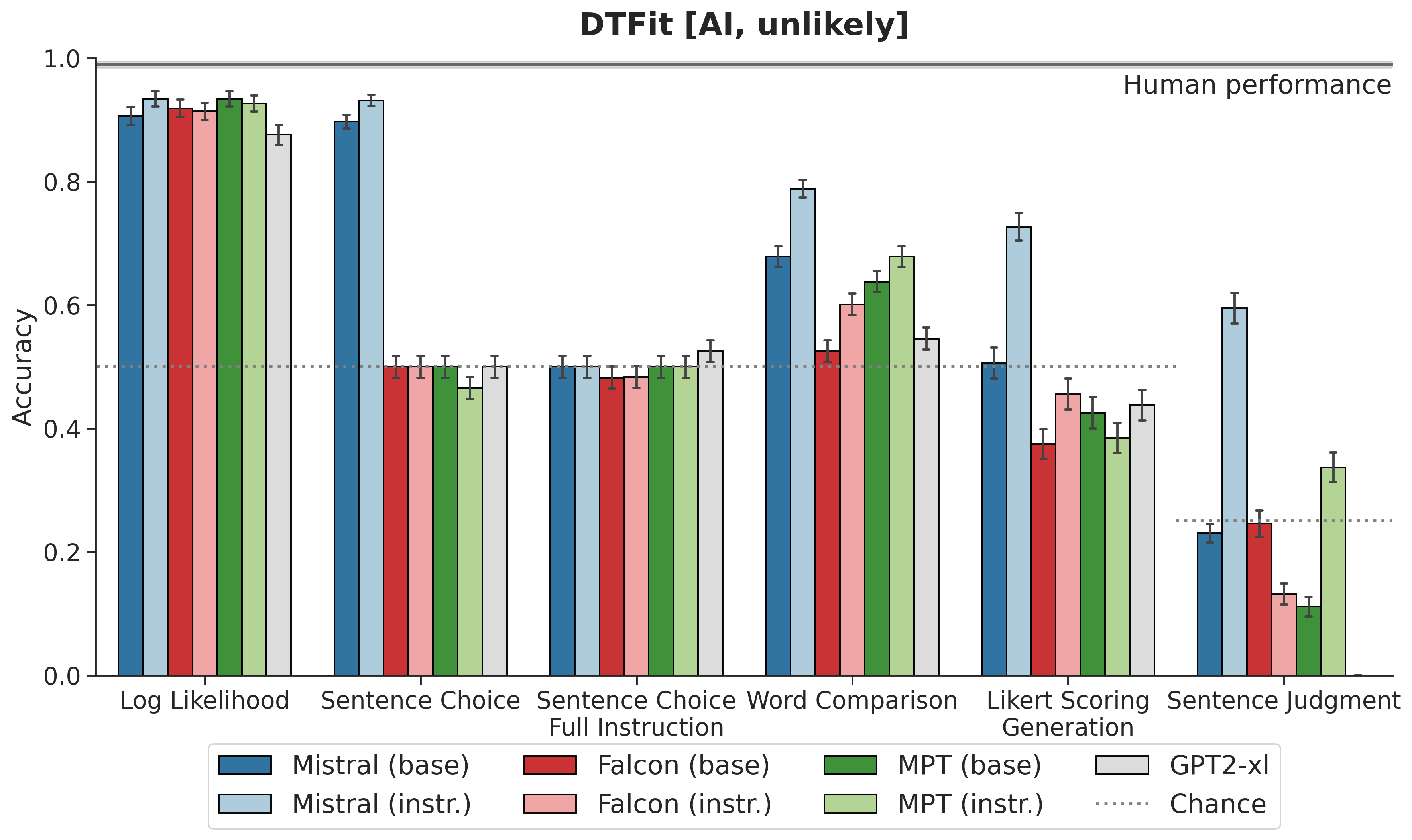}}
\captionof{figure}{Prompting results for DTFit, including best prompt from \citet{hu2023prompting}.}%
\label{fig:DTFit-prompting}
\end{minipage}

\section{Evidence for invariance to prompting variations for DTFit}\label{sec:si-invariance-prompting}

\subsection{Free vs. constrained generation}

Here, we evaluate prompt-based generation in two ways: using a free vs. constrained generation paradigm. In the free paradigm, we ask the model to generate up to $20$ tokens in the completion and find responses that include a valid response (exactly one numeral between 1-2 or 1-7). In the constrained paradigm, we only allow completions from a predefined set of tokens, i.e., either the set \{1,2\} or the set \{1,2,3,4,5,6,7\}, using a regex-matching generation procedure from {\tt outlines}\footnote{\url{https://github.com/outlines-dev/outlines}}. Results are roughly consistent across metrics, yielding no advantage of one over the other prompting paradigm in both \textit{Sentence Choice} and \textit{Likert Scoring} paradigms.

\vspace*{0.1em}
\noindent%
\begin{minipage}{\linewidth}
\centering
\makebox[\linewidth]{
\includegraphics[width=\textwidth]{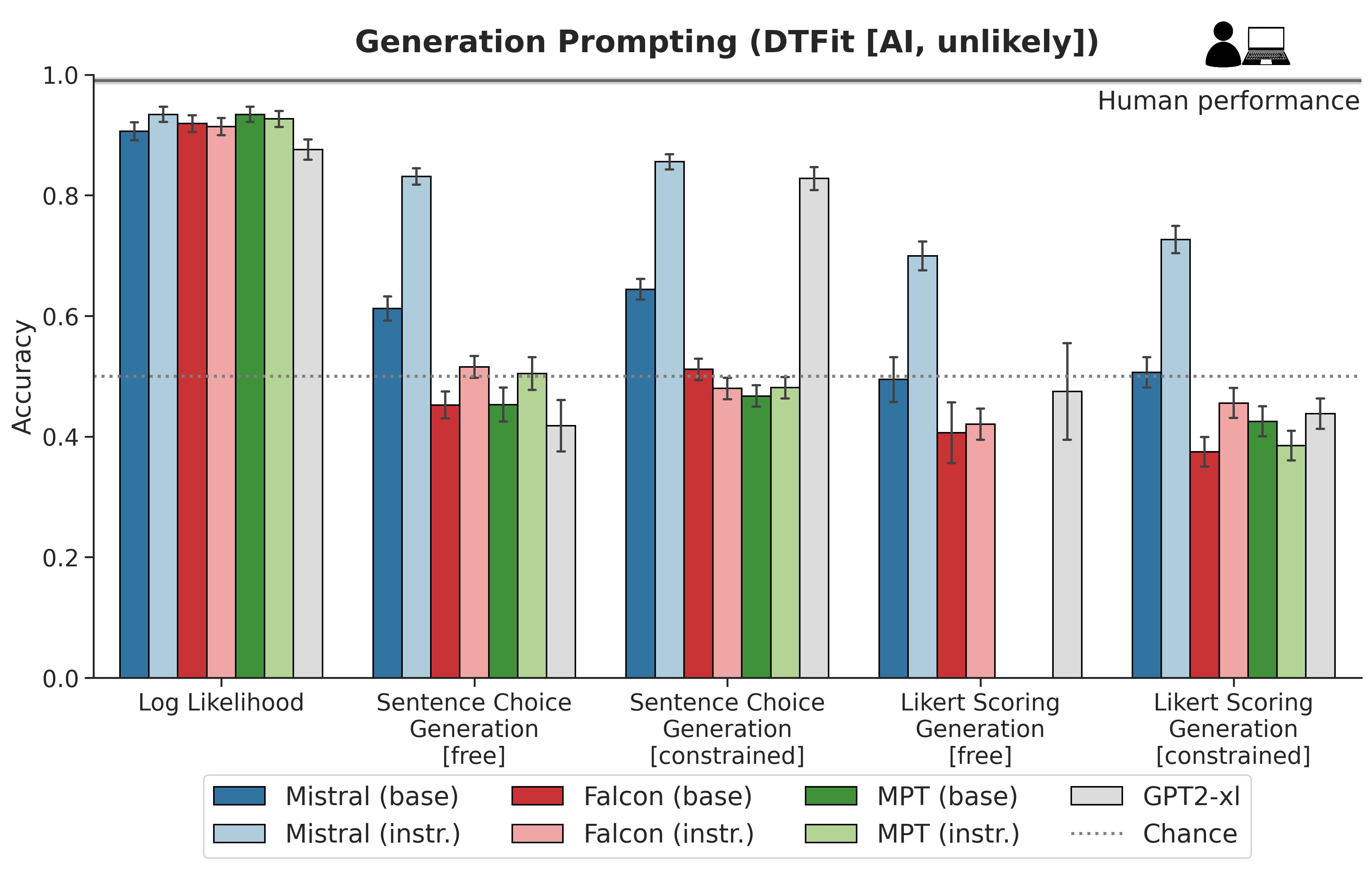}}
\captionof{figure}{Comparison of free vs. constrained generation prompting. Note that {\tt MPT} results are missing for the free Likert Scoring method.}%
\label{fig:free-vs-constrained}
\end{minipage}
\vspace*{0.1em}

\subsection{Query types}

\vspace*{0.1em}
\noindent%
\begin{minipage}{\linewidth}
\centering
\makebox[\linewidth]{
\includegraphics[width=\textwidth]{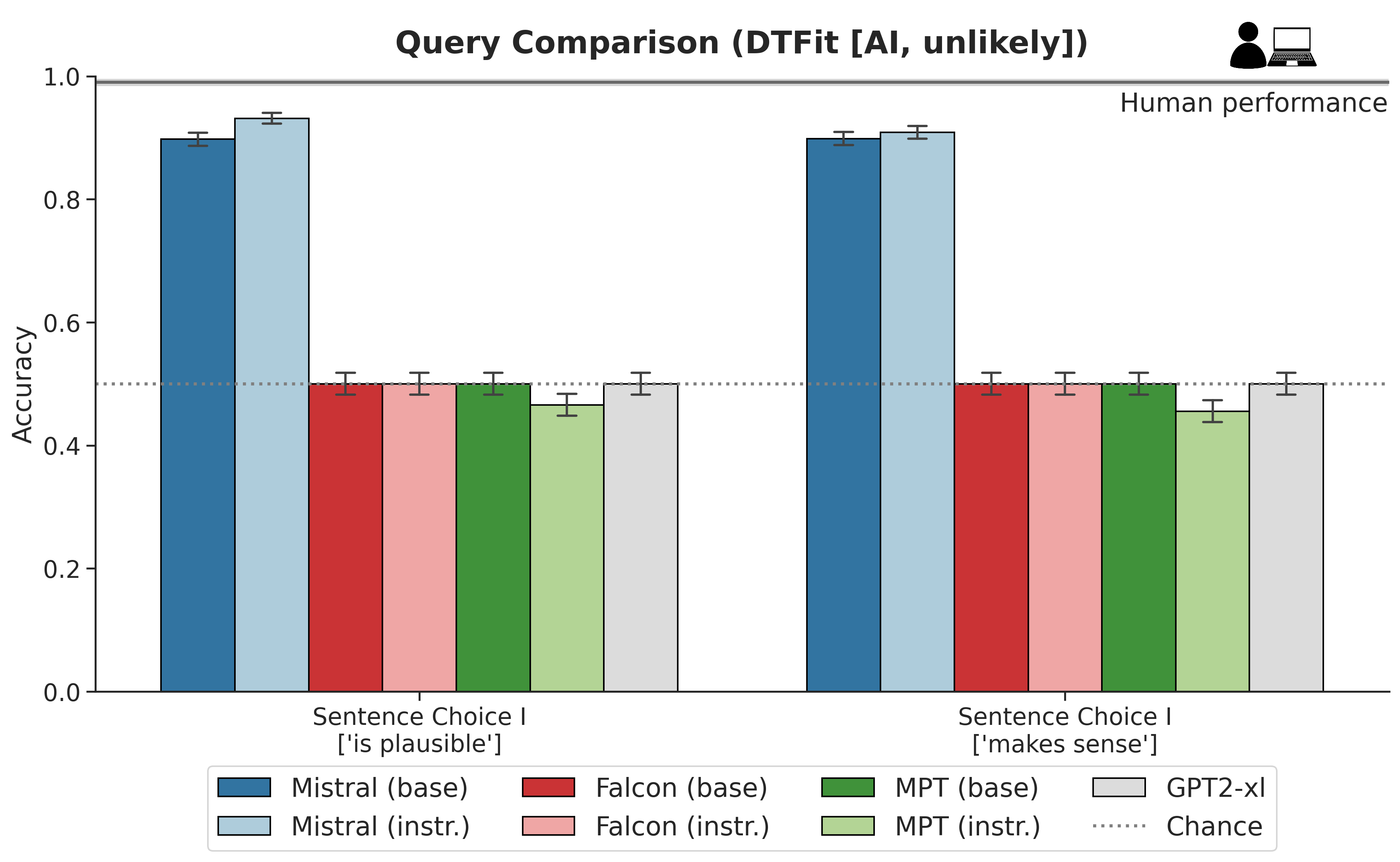}}
\captionof{figure}{Comparison of different query types for prompts of type \textit{Sentence Choice I}. In all supplementary figures for Experiment 1, we also include {\tt GPT2-xl} as a baseline model.}%
\label{fig:makessense-vs-isplausible}
\end{minipage}
\vspace*{0.1em}

\section{GLMM analysis}\label{sec:glmm}
We fit a binomial generalized linear mixed-effects model (GLMM) with a logit link function to predict the binary variable accuracy, using LLM model class ({\tt Mistral}, {\tt Falcon}, {\tt MPT}), model version ({\tt base}, {\tt instruct}), and voice (active, passive) as fixed effects, and items as random effects. The model further included all interactions between the fixed effects. We used dummy coding for voice, with ``active'' as the reference level, and sum-coding for model class and model version. The analysis was conducted using the {\tt lme4} R package \citep{bates2014fitting}.

\section{Quantifying the fit to human result patterns for Experiment 2: Context-Dependent Plausibility Judgments}\label{sec:stats-socialN400}

To compare the result patterns of humans vs. models for the sentence sensibility judgment task across conditions and across both continuous (\logprobs) vs. discrete (\prompting) outputs (which for some items led to zero-variance response vectors across experimental conditions), we measured the similarity between human and model responses across different experimental conditions using Euclidean distance with the following approach. First, we scaled the response data for each model using  min-max scaling to prevent distance calculations to be biased by differences in response magnitude. For each pair of human and model responses, we then calculated the Euclidean distance between the three-point response vectors across conditions (Control, Critical, SemAnom) for each item. To convert this distance into a similarity value, we used a normalized metric where similarity is defined as $1 - \frac{\text{distance}}{\text{max distance}}$ where the maximum possible Euclidean distance between two vectors corresponds to the vector's dimensionality, yielding a similarity score in the range from $0$ (maximally dissimilar) to $1$ (identical). Similarity scores were calculated for all combinations of context (human context vs. model context, human context vs. model no context, human no context vs. model context, human no context vs. model no context). The similarity scores were then averaged across items to obtain a final similarity value for each of the four conditions. We report the average similarity scores per model across the matched (human and model both in ``Context'' or both in ``No Context'') and mismatched (one in ``Context'' and the other in ``No Context'') conditions in Tables \ref{tab:stats-socialn400-ll}, \ref{tab:stats-socialn400-judge}.

\begin{table}[t!]
\centering
\resizebox{0.9\columnwidth}{!}{%
\begin{tabular}{lcc}
\toprule
\textbf{Model} & \textbf{matched} & \textbf{unmatched} \\
\midrule
{\tt Mistral (base)} & 0.41 & 0.31 \\
{\tt Mistral (instruct)} & 0.40 & 0.30 \\
{\tt Falcon (base)} & 0.51 & 0.39 \\
{\tt Falcon (instruct)} & 0.51 & 0.40 \\
{\tt MPT (base)} & 0.50 & 0.38 \\
{\tt MPT (instruct)} & 0.48 & 0.37 \\
\bottomrule
\end{tabular}
}
\caption{Similarity results of human to model response pattern analysis for Figure \ref{fig:jouravlev2019}.}
\label{tab:stats-socialn400-ll}
\end{table}

\begin{table}[t!]
\centering
\resizebox{0.9\columnwidth}{!}{%
\begin{tabular}{lcc}
\toprule
\textbf{Model} & \textbf{matched} & \textbf{unmatched} \\
\midrule
{\tt Mistral (base)} & 0.06 & 0.08 \\
{\tt Mistral (instruct)} & 0.30 & 0.14 \\
{\tt Falcon (base)} & 0.04 & 0.04 \\
{\tt Falcon (instruct)} & 0.22 & 0.08 \\
{\tt MPT (base)} & -0.05 & -0.05 \\
{\tt MPT (instruct)} & 0.13 & 0.07 \\
\bottomrule
\end{tabular}
}
\caption{Similarity results of human to model response pattern analysis for Figure \ref{fig:jouravlev2019-judge}.}
\label{tab:stats-socialn400-judge}
\end{table}

We further conducted paired t-tests to compare similarity scores in matched context conditions with mismatched conditions in order to determine whether the models captured the human responses significantly better when the context matched. T-test results are reported in Tables \ref{tab:stats-socialn400-ll-ttest}, \ref{tab:stats-socialn400-judge-ttest}.

\begin{table}[b!]
\centering
\resizebox{0.9\columnwidth}{!}{%
\begin{tabular}{lcc}
\toprule
\textbf{Model} & \textbf{t-statistic} & \textbf{p-value} \\
\midrule
{\tt Mistral (base)} & 8.49 & 0.00 \\
{\tt Mistral (instruct)} & 8.52 & 0.00 \\
{\tt Falcon (base)} & 10.83 & 0.00 \\
{\tt Falcon (instruct)} & 9.69 & 0.00 \\
{\tt MPT (base)} & 11.80 & 0.00 \\
{\tt MPT (instruct)} & 10.70 & 0.00 \\
\bottomrule
\end{tabular}
}
\caption{T-test results to compare similarity scores in matched context conditions with mismatched conditions in Figure \ref{fig:jouravlev2019}.}
\label{tab:stats-socialn400-ll-ttest}
\end{table}

\begin{table}[t!]
\centering
\resizebox{0.9\columnwidth}{!}{%
\begin{tabular}{lcc}
\toprule
\textbf{Model} & \textbf{t-statistic} & \textbf{p-value} \\
\midrule
{\tt Mistral (base)} & -1.43 & 0.00 \\
{\tt Mistral (instruct)} & 4.81 & 0.15 \\
{\tt Falcon (base)} & 0.04 & 0.97 \\
{\tt Falcon (instruct)} & 4.69 & 0.00 \\
{\tt MPT (base)} & 0.01 & 0.99 \\
{\tt MPT (instruct)} & 2.84 & 0.01 \\
\bottomrule
\end{tabular}
}
\caption{T-test results to compare similarity scores in matched context conditions with mismatched conditions in Figure \ref{fig:jouravlev2019-judge}.}
\label{tab:stats-socialn400-judge-ttest}
\end{table}

\section{Replicating the sensibility-judgment task by \citet{jouravlev2019tracking} using prompting}\label{sec:si-jouralev-prompting}
To replicate the human experiment by \citet{jouravlev2019tracking} in LMs using prompting, we queried the models using an adjusted \textit{Sentence Judgment} prompt (see Table \ref{tab:prompts}): {[}No context:{]} \textit{Here is a sentence: ``{sentence}''. Does this sentence make sense? Respond with either Yes or No as your answer.} {[}With context:{]} \textit{Here is a context: ``{context}'', and here is a sentence: ``{sentence}''. Does this sentence make sense considering the context? Respond with either Yes or No as your answer.} We report our results in Figure \ref{fig:jouravlev2019-judge}.

We observe that while most base models often favor one answer option, the instruction-tuned models exhibit more a nuanced behavior: These models are more consistent with human responses in this binary sensitivity judgment task, matching them qualitatively. Nevertheless, instruction-tuned models tend to (i) systematically underestimate the contextual plausibility of the {\tt Critical} sentences (Figure \ref{fig:jouravlev2019-judge}, upper panel), and (ii) systematically overestimate the plausibility of implausible sentences relative to humans ({\tt SemAnom} conditions and {\tt Critical} condition, Figure \ref{fig:jouravlev2019-judge}, lower panel) in the binary sensibility-judgment task setup.

\section{Replicating the sensibility-judgment task by \citet{jouravlev2019tracking} using sentence log likelihoods}\label{sec:si-jouralev-sentence-ll}
\begin{figure*}[t!]
    \centering
    \includegraphics[width=\textwidth]{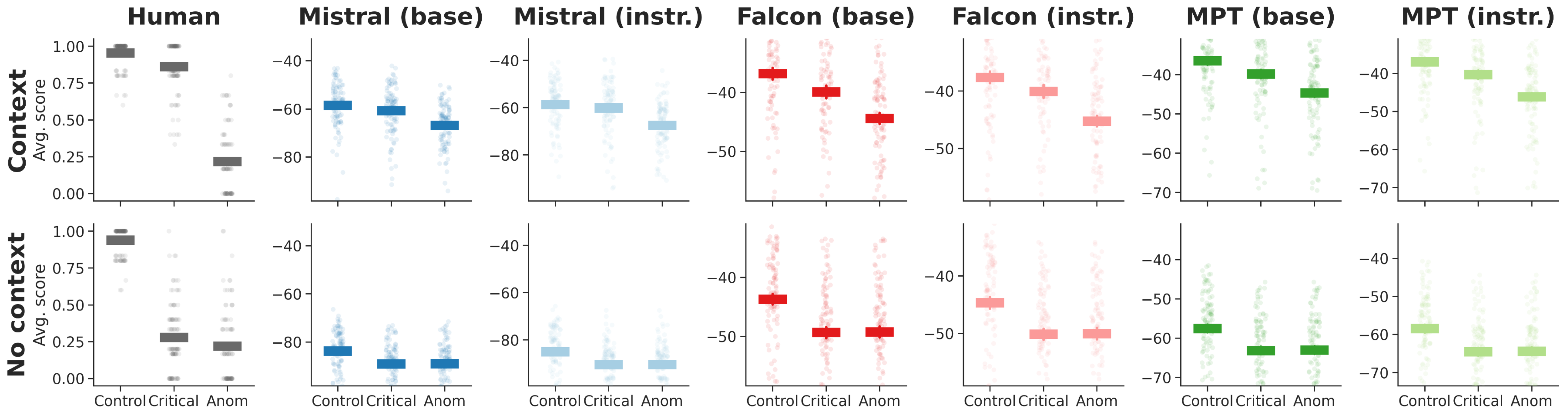}
    \caption{Replicating the sensibility-judgment task in LMs using sentence \logprobs\ measures. Human data from \citet{jouravlev2019tracking}.} 
    \label{fig:jouravlev2019-sentence-ll}
\end{figure*}

In Figure \ref{fig:jouravlev2019-sentence-ll}, we replicate the human experiment by \citet{jouravlev2019tracking} in LMs using sentence log likelihood measurements. We generally observe similar trends than the comparison with the target word measurement.

\end{document}